\documentclass[11pt,a4paper]{article}
\usepackage[hyperref]{acl2019}
\usepackage{times}
\usepackage{latexsym}

\usepackage{url}

\usepackage{amsmath}
\usepackage{amssymb}
\usepackage[utf8]{inputenc}
\usepackage{bbm}
\usepackage{graphicx}
\usepackage{multirow}

\usepackage{subcaption}

\usepackage[draft]{todonotes}

\usepackage[LGR,T1]{fontenc}

\DeclareSymbolFont{upgreek}{LGR}{cmr}{m}{n}
\DeclareMathSymbol{\upsigma}{\mathord}{upgreek}{`s}

\DeclareMathOperator{\softmax}{softmax}
\DeclareMathOperator{\encoder}{EncoderRNN}
\DeclareMathOperator{\decoder}{DecoderRNN}
\DeclareMathOperator{\outpt}{Output}
\DeclareMathOperator{\energy}{Energy}
\DeclareMathOperator{\monotonicenergy}{MonotonicEnergy}
\DeclareMathOperator{\softmaxenergy}{SoftmaxEnergy}
\DeclareMathOperator{\bernoulli}{Bernoulli}

\newcommand*\samethanks[1][\value{footnote}]{\footnotemark[#1]}

\aclfinalcopy 
\def\aclpaperid{2448} 

\title{Monotonic Infinite Lookback Attention\\
for Simultaneous Machine Translation}

\author{Naveen Arivazhagan\thanks{~~Equal contributions.}
\And Colin Cherry\samethanks[1] \And Wolfgang Macherey\And Chung-Cheng Chiu
\AND Semih Yavuz\\\\
~~~~~~~~~~~~~~~~~~~~~~~~~~~~~~~~~~~~~~~~~~~~~~~~~~~~~~~~~~~~~~~~~~~~~~~~~~~~~~~~~~~~~~~~~~~~~~~~~~~~~~~~~~~~~~~~~~~~~~~Google\\
~~~~~~~~~~~~~~~~~~~~~~~~~~~~~~~~~~~~~~~~~~~~~~~~~~~~~~~~~~~~~~~~~~~~~~~~~~~~~~~~~~~~~~~~~~~~~~~~~~~~~~~~~~~~~~~~~~~~~\texttt{navari,colincherry,wmach,chungchengc@google.com}\\
~~~~~~~~~~~~~~~~~~~~~~~~~~~~~~~~~~~~~~~~~~~~~~~~~~~~~~~~~~~~~~~~~~~~~~~~~~~~~~~~~~~~~~~~~~~~~~~~~~~~~~~~~~~~~~~~~~~~~\texttt{syavuz,rpang,mweili,craffel@google.com}\\
    \And Ruoming Pang \And Wei Li \And Colin Raffel
}
\date{}

\begin{document}
\maketitle
\begin{abstract}
Simultaneous machine translation begins to translate each source sentence before the source speaker is finished speaking, with applications to live and streaming scenarios. Simultaneous systems must carefully schedule their reading of the source sentence to balance quality against latency. We present the first simultaneous translation system to learn an adaptive schedule jointly with a neural machine translation (NMT) model that attends over all source tokens read thus far. 
We do so by introducing Monotonic Infinite Lookback (MILk) attention, which maintains both a hard, monotonic attention head to schedule the reading of the source sentence, and a soft attention head that extends from the monotonic head back to the beginning of the source. 
We show that MILk's adaptive schedule allows it to arrive at latency-quality trade-offs that are favorable to those of a recently proposed \mbox{wait-$k$} strategy for many latency values.
\end{abstract}

\newcommand\size[1]{|\mathbf{#1}|}
\newcommand\ALy[0]{\mathrm{AL}_{\size{y}}}
\newcommand\AL[0]{\mathrm{AL}}
\newcommand\AFL[0]{\mathrm{AFL}}
\newcommand\ex[1]{\textit{#1}}

\section{Introduction}
Simultaneous machine translation (MT) addresses the problem of how to begin translating 
a source sentence before the source speaker has finished speaking.
This capability is crucial for live or streaming translation scenarios, such as 
speech-to-speech translation, where waiting for one speaker to complete their sentence 
before beginning the translation would introduce an intolerable delay.
In these scenarios, the MT engine must balance latency against quality:
if it acts before the necessary source content arrives, translation quality degrades;
but waiting for too much source content can introduce unnecessary delays.
We refer to the strategy an MT engine uses to balance reading source tokens 
against writing target tokens as its \textbf{schedule}.

Recent work in simultaneous machine translation tends to fall into one of two bins:
\begin{itemize}
    \item The schedule is learned and/or adaptive to the current context, but assumes
    a fixed MT system trained on complete source sentences, as typified by 
    wait-if-*~\cite{Cho16} and reinforcement learning 
    approaches~\cite{Grissom2014,Gu2017}.
    \item The schedule is simple and fixed and can thus be easily integrated into MT training, as typified
    by wait-$k$ approaches~\cite{Dalvi2018,Ma2018}.
\end{itemize}
Neither scenario is optimal.
A fixed schedule may introduce too much delay for some sentences, and not
enough for others.
Meanwhile, a fixed MT system that was trained to expect complete 
sentences may impose a low ceiling on any adaptive schedule that uses it.
Therefore, we propose to train an adaptive schedule jointly with the underlying neural machine
translation (NMT) system.

Monotonic attention mechanisms~\cite{Raffel2017,Chiu2018} are designed for integrated training in streaming scenarios and provide our starting point. They encourage streaming by confining the scope of attention to the most recently read tokens. This restriction, however, may hamper long-distance reorderings that can occur in MT. We develop an approach that removes this limitation while preserving the ability to stream.

We use their hard, monotonic attention head to determine how much
of the source sentence is available. Before writing each target token, our learned model advances this head zero
or more times based on the current context, with each advancement
revealing an additional token of the source sentence. A secondary, soft attention head can then attend to any source words at or before that 
point, resulting in \textbf{M}onotonic \textbf{I}nfinite \textbf{L}ookbac\textbf{k} 
(MILk) attention. This, however, removes the memory constraint that was encouraging the model to stream. To restore streaming behaviour, we propose to jointly minimize a latency loss. The entire system can efficiently be trained in expectation, as a drop-in replacement for the familiar soft attention.

%
%
Our contributions are as follows:
\begin{enumerate}
    \item We present MILk attention, which allows us to build
     the first  simultaneous  MT  system  to  learn an adaptive schedule jointly with an NMT model  that  attends over all source tokens read thus far.
    \item We extend the recently-proposed Average Lagging 
    latency metric~\cite{Ma2018}, making it differentiable and calculable in 
    expectation, which allows it to be used as a training objective.
    \item We demonstrate favorable trade-offs to those of wait-$k$ strategies at many latency values, and provide evidence that MILk's advantage extends from its ability to adapt based on source content.
\end{enumerate}

\begin{figure*}[t]
    \centering
    \begin{subfigure}[b]{0.3\textwidth}
        \includegraphics[width=\textwidth]{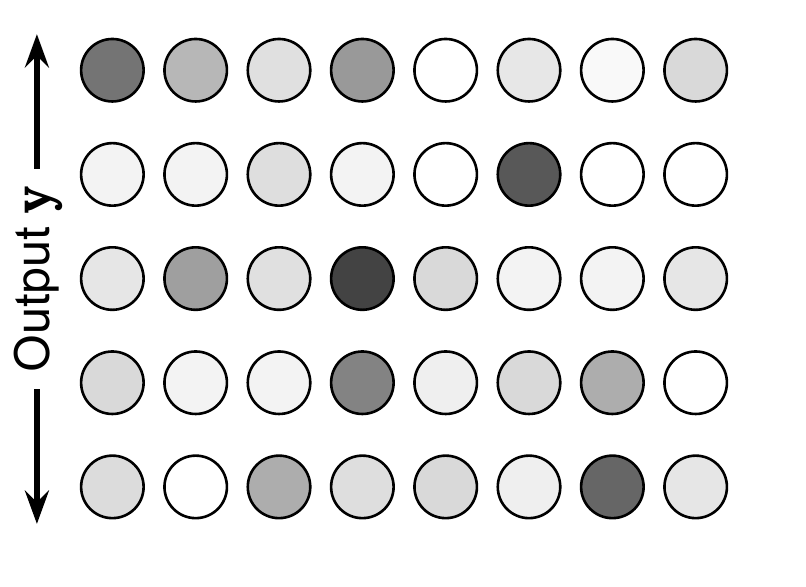}
        \caption{Soft attention.}
        \label{fig:softmax_grid}
    \end{subfigure}
    \begin{subfigure}[b]{0.3\textwidth}
        \includegraphics[width=\textwidth]{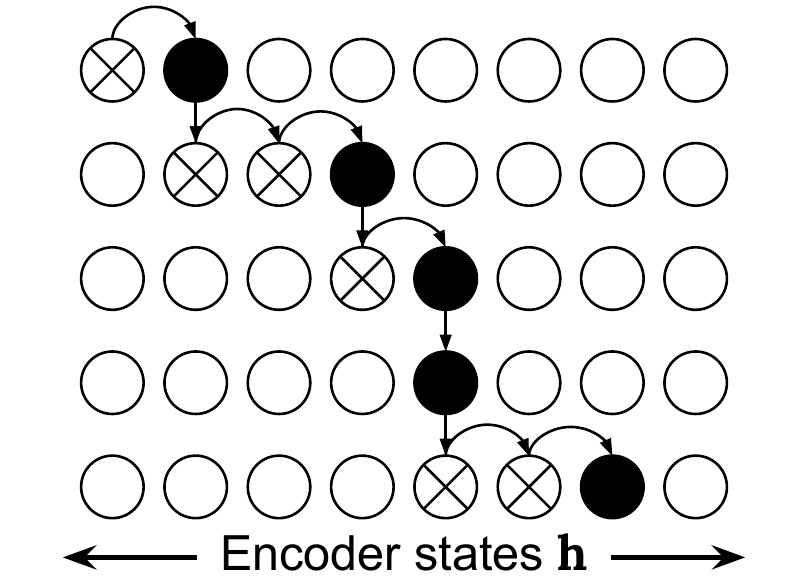}
        \caption{Monotonic attention.}
        \label{fig:monotonic_grid}
    \end{subfigure}
    \begin{subfigure}[b]{0.3\textwidth}
        \includegraphics[width=\textwidth]{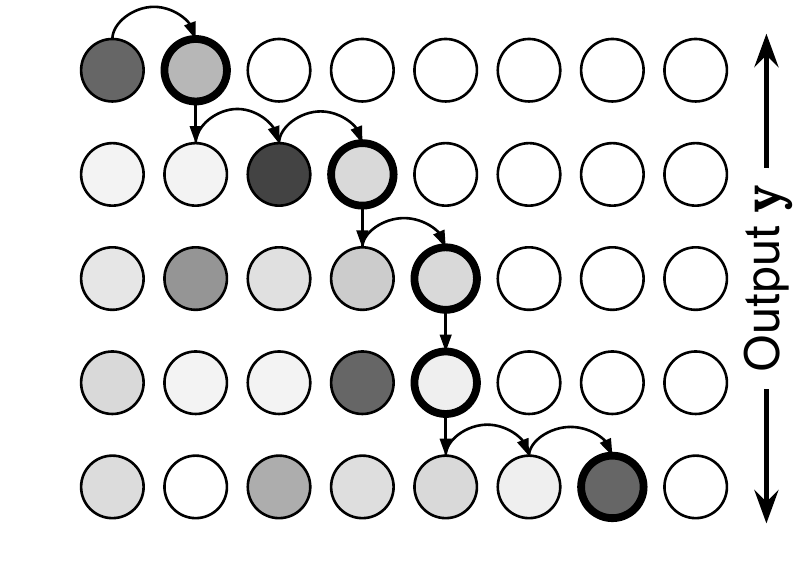}
        \caption{MILk attention.}
        \label{fig:mocha_grid}
    \end{subfigure}
    \caption{
    Simplified diagrams of the attention mechanisms discussed in Sections \ref{sec:Monotonic} and \ref{sec:Milk}.
    The shading of each node indicates the amount of attention weight the model assigns to a given encoder state (horizontal axis) at a given output timestep (vertical axis).
    }
    \label{fig:attention_grids}
\end{figure*}

\section{Background}
Much of the earlier work on simultaneous MT took the form
of strategies to chunk the source sentence into partial segments that can be translated safely. These segments could be triggered by prosody~\cite{Fugen2007,Bangalore2012} or lexical cues~\cite{RangarajanSridhar2013},
or optimized directly for translation quality~\cite{Oda2014}.
Segmentation decisions are surrogates for the core problem, which is deciding whether enough source content has been read to write the next target word correctly~\cite{Grissom2014}. However, since doing so involves discrete decisions, learning via back-propagation is obstructed. Previous work on simultaneous NMT has thus far side-stepped this problem by making restrictive simplifications, either on the underlying NMT model or on the flexibility of the schedule.

\newcite{Cho16} apply heuristics measures to estimate and then threshold the confidence of an NMT model trained on full sentences to adapt it at inference time to the streaming scenario. Several others use reinforcement learning (RL) to develop an agent to predict read and write decisions~\cite{Satija2016,Gu2017,Alinejad2018}. However, due to computational challenges, they pre-train an NMT model on full sentences and then train an agent that sees the fixed NMT model as part of its environment.

\newcite{Dalvi2018} and \newcite{Ma2018} use fixed schedules and train their NMT systems accordingly. In particular, \newcite{Ma2018} advocate for a wait-$k$ strategy, wherein the system always waits for exactly $k$ tokens before beginning to translate, and then alternates between reading and writing at a constant pre-specified emission rate. Due to the deterministic nature of their schedule, they 
can easily train the NMT system with the schedule in place.
This can allow the NMT system to learn to anticipate missing content using its
inherent language modeling capabilities.
On the downside, with a fixed schedule the model cannot speed up or slow down appropriately for particular inputs. 

\newcite{Press2018} recently developed an attention-free model that aims to reduce computational and memory requirements. They achieve this by maintaining a single running context vector, and eagerly emitting target tokens based on it  whenever possible. Their method is adaptive and uses integrated training, but the schedule itself is trained with external supervision provided by word alignments, while ours is latent and learned in service to the MT task. 

\section{Methods}
\label{sec:methods}

In sequence-to-sequence modeling, the goal is to transform an input sequence $\mathbf{x} = \{x_1, \ldots, x_{\size{x}}\}$ into an output sequence $\mathbf{y} = \{y_1, \ldots, y_{\size{y}}\}$.
A sequence-to-sequence model consists of an encoder which maps the input sequence to a sequence of hidden states and a decoder which conditions on the encoder output and autoregressively produces the output sequence.
In this work, we consider sequence-to-sequence models where the encoder and decoder are both recurrent neural networks (RNNs) and are updated as follows:
\begin{align}
    h_j &= \encoder(x_j, h_{j - 1}) \\
    s_i &= \decoder(y_{i - 1}, s_{i - 1}, c_i) \\
    y_i &= \outpt(s_i, c_i)
\end{align}
where $h_j$ is the encoder state at input timestep $j$, $s_i$ is the decoder state at output timestep $i$, and $c_i$ is a context vector.
The context vector is computed based on the encoder hidden states through the use of an attention mechanism \cite{bahdanau2014neural}.
The function $\outpt(\cdot)$ produces a distribution over output tokens $y_i$ given the current state $s_i$ and context vector $c_i$.
In standard soft attention, the context vector is computed as follows:
\begin{align}
    \label{eqn:energy} e_{i, j} &= \energy(h_j, s_{i - 1}) \\
    \alpha_{i, j} &= \softmax(e_{i, :})_j := \frac{\exp(e_{i, j})}{\sum_{k = 1}^T \exp(e_{i, k})}\\
    c_i &= \sum_{j = 1}^{\size{x}} \alpha_{i, j}h_j \label{eq:ci_soft}
\end{align}
where $\energy()$ is a multi-layer perceptron.

One issue with standard soft attention is that it computes $c_i$ based on the entire input sequence for all output timesteps; this prevents attention from being used in streaming settings since the entire input sequence needs to be ingested before generating any output. To enable streaming, we require a schedule in which the output at timestep $i$ is generated using just the first $t_i$ input tokens, where $1 \leq t_i \leq \size{x}$.

\subsection{Monotonic Attention}
\label{sec:Monotonic}
\newcite{Raffel2017} proposed a monotonic attention mechanism that modifies standard soft attention to provide such a schedule of interleaved reads and writes, while also integrating training with the rest of the NMT model. Monotonic attention explicitly processes the input sequence in a left-to-right order and makes a hard assignment of $c_i$ to one particular encoder state denoted $h_{t_i}$.
For output timestep $i$, the mechanism begins scanning the encoder states starting at $j = t_{i - 1}$. For each encoder state, it produces a Bernoulli selection probability $p_{i, j}$, which corresponds to the probability of either stopping and setting $t_i = j$, or else moving on to the next input timestep, $j+1$, which represents reading one more source token.
This selection probability is computed through the use of an energy function that is passed through a logistic sigmoid to parameterize the Bernoulli random variable:
\begin{align}
        e_{i, j} &= \monotonicenergy(s_{i - 1}, h_j)\\
        p_{i, j} &= \upsigma(e_{i, j}) \label{eq:pij} \\
        z_{i, j} &\sim \bernoulli(p_{i, j})
\end{align}
If $z_{i, j} = 0$, $j$ is incremented and these steps are repeated; if $z_{i, j} = 1$, $t_i$ is set to $j$ and $c_i$ is set to $h_{t_i}$.

This approach involves sampling a discrete random variable and a hard assignment of \mbox{$c_i = h_{t_i}$}, which precludes backpropagation.
\newcite{Raffel2017} instead compute the probability that \mbox{$c_i = h_j$} and use this to compute the expected value of $c_i$, which can be used as a drop-in replacement for standard soft attention, and which allows for training with backpropagation.
The probability that the attention mechanism attends to state $h_j$ at output timestep $i$ is computed as
\begin{equation}
    \alpha_{i, j} = p_{i, j} \left((1 - p_{i, j - 1})\frac{\alpha_{i, j - 1}}{p_{i, j - 1}} + \alpha_{i - 1, j}\right)
\end{equation}
There is a solution to this recurrence relation which allows $\alpha_{i, j}$ to be computed for all $j$ in parallel using cumulative sum and cumulative pro\-duct operations; see \newcite{Raffel2017} for details. 

Note that when $p_{i, j}$ is either $0$ or $1$, the soft and hard approaches are the same.
To encourage this, \newcite{Raffel2017} use the common approach of adding zero-mean Gaussian noise to the logistic sigmoid function's activations.
Equation~\ref{eq:pij} becomes:
\begin{equation}
    p_{i, j} = \upsigma\left(e_{i, j} + \mathcal{N}(0, n)\right) \label{eq:psn}
\end{equation}
One can control the extent to which $p_{i,j}$ is drawn toward discrete values
by adjusting the noise variance $n$.
At run time, we forgo sampling in favor of simply setting $z_{i,j} = \mathbbm{1}{(e_{i,j} > 0 )}$.

While the monotonic attention mechanism allows for streaming attention, it requires that the decoder attend only to a single encoder state, $h_{t_i}$.
To address this issue, \newcite{Chiu2018} proposed monotonic chunkwise attention (MoChA), which allows the model to perform soft attention over a small fixed-length chunk preceding $t_i$, i.e.\ over all available encoder states, $h_{t_i - \mathit{cs} + 1}, h_{t_i - \mathit{cs} + 2}, \ldots, h_{t_i}$ for some fixed chunk size $\mathit{cs}$.

\subsection{Monotonic Infinite Lookback Attention}
\label{sec:Milk}
In this work, we take MoChA one step further, by allowing the model to perform soft attention over the encoder states $h_1, h_2, \ldots, h_{t_i}$.
This gives the model ``infinite lookback'' over the past seen thus far, so we dub this technique \textbf{M}onotonic \textbf{I}nfinite \textbf{L}ookbac\textbf{k} (MILk) attention.
The infinite lookback provides more flexibility and should improve the modeling of
long-distance reorderings and dependencies. The increased computational cost, from linear to quadratic computation, is of little concern as our focus on the simultaneous scenario means that out largest source of latency will be waiting for source context.

Concretely, we maintain a full monotonic attention mechanism
and also a soft attention mechanism.
Assuming that the monotonic attention component chooses to stop at $t_i$, MILk first computes soft attention energies
\begin{equation}
    u_{i, k} = \softmaxenergy(h_k, s_{i - 1})
\end{equation}
for $k \in {1, 2, \ldots, t_i}$ where $\softmaxenergy(\cdot)$ is an energy function similar to Equation (\ref{eqn:energy}). 
Then, MILk computes a context $c_i$ by
\begin{equation}
    c_i = \sum_{j = 1}^{t_i} \frac{\exp(u_{i, j})}{\sum_{l = 1}^{t_i} \exp(u_{i, l})}h_j
\end{equation}
Note that a potential issue with this approach is that the model can set the monotonic attention head $t_i = \size{x}$ for all $i$, in which case the approach is equivalent to standard soft attention.
We address this issue in the following subsection.

To train models using MILk, we compute the expected value of $c_i$ given the monotonic attention probabilities and soft attention energies.
To do so, we must consider every possible path through which the model could assign attention to a given encoder state.
Specifically, we can compute the attention distribution induced by MILk by
\begin{equation} \label{eqn:beta}
    \beta_{i, j} = \sum_{k = j}^{\size{x}} \left( \frac{\alpha_{i, k}\exp(u_{i, j})}{\sum_{l = 1}^k \exp(u_{i, l})} \right)
\end{equation}
The first summation reflects the fact that $h_j$ can influence $c_i$ as long as 
$k \geq j$, 
and the term inside the summation reflects the attention probability associated with some monotonic probability $\alpha_{i, k}$ and the soft attention distribution.
This calculation can be computed efficiently using cumulative sum operations by replacing the outer summation with a cumulative sum and the inner operation with a cumulative sum after reversing $\mathbf{u}$.
Once we have the $\beta_{i,j}$ distribution, calculating the expected context $c_i$ follows a familiar formula: $c_i=\sum_{j=1}^{\size{x}} \beta_{i,j} h_j$.

\subsection{Latency-augmented Training}
\label{sec:LatencyAugmented}
By moving to an infinite lookback, we have gained the full power of a soft attention mechanism
over any source tokens that have been revealed up to time $t_i$.
However, while the original monotonic attention encouraged streaming behaviour implicitly due to the restriction on the system's memory, MILk no longer has any incentive to do this. It can simply wait for all source tokens before writing the first target token.
We address this problem by training with an objective that interpolates log likelihood with a
latency metric.

Sequence-to-sequence models are typically trained to minimize the negative log likelihood,
which we can easily augment with a latency cost:
\begin{equation}
\label{eq:obj}
    L(\theta) = -\sum_{(\mathbf{x}, \mathbf{y})}\log p(\mathbf{y}|\mathbf{x};\theta) 
    + \lambda \mathcal{C}(\mathbf{g})
\end{equation}
where 
$\lambda$ is a user-defined latency weight,
$\mathbf{g} = \{g_1, \ldots, g_{\size{y}}\}$  is a vector that describes the delay incurred immediately before each target time step 
(see Section~\ref{sec:LatencyMetrics}),
and $\mathcal{C}$ is a latency metric that transforms these delays into a cost.

In the case of MILk, $g_i$ is equal to $t_i$, the position of the monotonic attention 
head.\footnote{We introduce $g_i$ to generalize beyond methods with hard attention heads and to unify notation with \newcite{Ma2018}.}
Recall that during training, we never actually make a hard decision about $t_i$'s location. Instead, we can use $\alpha_{i,j}$, the probability that $t_i=j$, to get expected delay:
\begin{equation}
    \label{eq:delay}
    g_i = \sum_{j=1}^{\size{x}} j\alpha_{i,j}
\end{equation}
So long as our metric is differentiable and well-defined over fractional delays,
Equation~(\ref{eq:obj}) can be used to guide MILk to low latencies.

\subsection{Preserving Monotonic Probability Mass}
\label{sec:residual_fix}
In the original formulations of monotonic attention (see Section~\ref{sec:Monotonic}), it is possible to choose not to stop the monotonic attention head, even at the end of the source sentence. In such cases, the attention returns an all-zero context vector. 

In early experiments, we found that this creates 
an implicit incentive for low latencies:
the MILk attention head would stop early to avoid running off the end of the sentence.
This implicit incentive grows stronger
as our selection probabilities $p_{i,j}$ come closer to being binary 
decisions.
Meanwhile, we found it beneficial to have very-near-to-binary decisions in order to get accurate latency estimates for latency-augmented training.
Taken all together, we found  that MILk either destabilized, or 
settled into unhealthily-low-latency regions.
We resolve this problem by forcing MILk's monotonic attention head to  once stop when it reaches the EOS token,  by setting $p_{i,\size{x}} = 1$.%
\footnote{While training, we perform the equivalent operation of shifting the any residual probability mass from overshooting the source sentence, $1 - \sum_{j=1}^{\size{x}} \alpha_{i,j}$, to the final source token at position $\size{x}$. This bypasses floating point errors introduced by the parallelized cumulative sum and cumulative product operations \cite{Raffel2017}. This same numerical instability helps explain why the parameterized stopping probability $p_{i,j}$ does not learn to detect the end of the sentence without intervention.}

\section{Measuring Latency}
\label{sec:IAL}
Our plan hinges on having a latency cost that is worth optimizing.
To that end, we describe two candidates, and then modify the
most promising one to accommodate our training scenario.

\subsection{Previous Latency Metrics}
\label{sec:LatencyMetrics}

\newcite{Cho16} introduced Average Proportion (AP), which
averages the absolute delay incurred by each target token:
\begin{equation}
  \mathrm{AP}=
  \frac{1}{\size{x}\phantom{\cdot}\size{y}}\sum_{i=1}^{\size{y}}g_i
\end{equation}
where $g_i$ is delay at time $i$:
the number of source tokens read by the agent
before writing the $i^{th}$ target token.
This metric has some nice properties, such as being bound between 0
and 1, but it also has some issues.
\newcite{Ma2018} observe that their \mbox{wait-$k$} system
with a fixed $k=1$ incurs different AP values as sequence
length $\size{x}=\size{y}$ ranges from 2 ($\mathrm{AP}=0.75$)
to $\infty$ ($\mathrm{AP}=0.5$).
Knowing that a very-low-latency wait-1 system incurs at best an
$\mathit{AP}$ of 0.5 also implies that much of the metric's dynamic
range is wasted; in fact, 
\newcite{Alinejad2018} report that AP is not sufficiently sensitive to
detect their improvements to simultaneous MT.

Recently, \newcite{Ma2018} introduced Average Lagging (AL),
which measures the average rate by which the MT system lags behind an
ideal, completely simultaneous translator:
\begin{equation}
  \label{eq:AL}
  \mathrm{AL} = \frac{1}{\tau}\sum_{i=1}^{\tau} g_i-\frac{i-1}{\gamma}
\end{equation}
where $\tau$ is the earliest timestep where the MT system has consumed
the entire source sequence:
\begin{equation}
\label{eq:tau}
\tau = \mathrm{argmin}_{i} g_i=\size{x}
\end{equation}
and $\gamma = \size{y}/\size{x}$ accounts for the source and
target having different sequence lengths.
This metric has the nice property that when $\size{x}=\size{y}$,
a wait-$k$ system will achieve an AL of $k$, which makes the metric
very interpretable.
It also has no issues with sentence length or sensitivity.

\subsection{Differentiable Average Lagging}
Average Proportion already works as a $\mathcal{C}$ function, 
but we prefer Average Lagging for the reasons outlined above.
Unfortunately, it is not differentiable, nor is it calculable in expectation, due to the $\mathrm{argmin}$ in Equation~(\ref{eq:tau}).
We present Differentiable Average Lagging (DAL), which eliminates the 
$\mathrm{argmin}$ by making AL's treatment of delay internally consistent.

AL's $\mathrm{argmin}$ is used to calculate $\tau$, which is used in turn to 
truncate AL's average at the point where all source tokens have been read.
Why is this necessary?
We can quickly see $\tau$'s purpose by reasoning about a simpler version of AL where
$\tau=\size{y}$.
\begin{table}[t]
  \begin{center}
  \begin{tabular}{rcccc||c|c}
    \multicolumn{5}{c||}{Statistics}&\multicolumn{2}{c}{Scores} \\ \hline
    $i$       & 1 & 2 & 3 & 4 & $\tau=2$ & $\tau=\size{y}$ \\
    $g_i$    & 3 & 4 & 4 & 4 &    &  \\
    $\mathrm{AL}_i$ & 3 & 3 & 2 & 1 & AL = 3  & AL = 2.25 \\
  \end{tabular}
  \end{center}
  \caption{
    Comparing AL with and without its truncated average, tracking 
    time-indexed lag $\mathrm{AL}_i=g_i-\frac{i-1}{\gamma}$ when $\size{x}=\size{y}=4$ 
    for a wait-$3$ system. \label{tab:exEqualLengths}}
\end{table}
Table~\ref{tab:exEqualLengths} shows the time-indexed lags that are averaged 
to calculate AL for a wait-3 system.
The lags make the problem clear: 
each position beyond the point where all source tokens have been read ($g_i=\size{x}$) 
has its lag reduced by 1, pulling the average lag below $k$.
By stopping its average at $\tau=2$, AL maintains the property that a wait-$k$ system receives an AL of $k$.

$\tau$ is necessary because the
only way to incur delay is to read a source token.
Once all source tokens have been read, all target tokens appear instantaneously, 
artificially dragging down the average lag.
This is unsatisfying: the system lagged behind the source speaker while they were
speaking. It should continue to do so after they finished.

AL solves this issue by truncating its average,
enforcing an implicit and poorly defined delay for the excluded, problematic tokens.
We propose instead to enforce a minimum delay for writing any target token.
Specifically, we model each target token as taking at least $\frac{1}{\gamma}$ units of
time to write, mirroring the speed of the ideal simultaneous translator
in AL's Equation~(\ref{eq:AL}).
We wrap $\mathbf{g}$ in a $\mathbf{g'}$ that enforces our minimum delay:
\begin{equation}
  g'_i = \left\{
  \begin{array}{ll}
    g_i & i=1\\
    \max\big(g_i, g'_{i-1}+\frac{1}{\gamma}\big) & i>1
  \end{array}\right.
\end{equation}
Like $g_i$, $g'_i$ represents the amount of delay incurred just before writing
the $i^{th}$ target token.
Intuitively, the $\max$ enforces our minimum delay:
$g'_i$ is either equal to $g_i$, the number of source tokens read,
or to $g'_{i-1}+\frac{1}{\gamma}$, the delay incurred just before the previous token,
plus the time spent writing that token.
The recurrence ensures that we never lose track of earlier delays.
With $\mathbf{g'}$ in place, we can define our Differentiable Average Lagging:
\begin{equation}
    \mathrm{DAL} = \frac{1}{\size{y}} \sum_{i=1}^{\size{y}} g'_i - \frac{i-1}{\gamma}
\end{equation}

\begin{table}[t]
  \begin{center}
  \begin{tabular}{rcccc||c}
    \multicolumn{5}{c||}{Statistics}&\multicolumn{1}{c}{Scores} \\ \hline
    $i$       & 1 & 2 & 3 & 4 &   \\
    $g'_i$    & 3 & 4 & 5 & 6 &   \\
    $\mathrm{DAL}_i$    & 3 & 3 & 3 & 3 & DAL = 3 \\
  \end{tabular}
  \end{center}
  \caption{
    DAL's
    time-indexed lag \mbox{$\mathrm{DAL}_i=g'_i-\frac{i-1}{\gamma}$} when $\size{x}=\size{y}=4$ 
    for a wait-$3$ system. \label{tab:IAL}}
\end{table}
DAL is equal to AL in many cases, in particular, when measuring wait-$k$ systems
for sentences of equal length, both always return a lag of $k$.
See Table~\ref{tab:IAL} for its treatment of our wait-3 example.
Having eliminated $\tau$, DAL is both differentiable and calcuable in
expectation.
\newcite{Cherry2019} provide further motivation and analysis for DAL, alongside
several examples of cases where DAL yields more intuitive results than AL.

\section{Experiments}

We run our experiments on the standard WMT14 English-to-French (EnFr; 36.3M sentences) and WMT15 German-to-English (DeEn; 4.5M sentences) tasks. For EnFr we use a combination of newstest 2012 and newstest 2013 for development and report results on newstest 2014. For DeEn we validate on newstest 2013 and then report results on newstest 2015. Translation quality is measured using detokenized, cased BLEU~\cite{Papineni2002}. For each data set, we use BPE~\cite{Sennrich2016} on the training data to construct a 32,000-type vocabulary that is shared between the source and target languages.

\subsection{Model}
Our model closely follows the RNMT+ architecture described by \newcite{Chen2018} with modifications to support streaming translation. It consists of a 6 layer LSTM encoder and an 8 layer LSTM decoder with additive attention~\cite{bahdanau2014neural}. All streaming models including wait-k, MoChA and MILk use unidirectional encoders, while offline translation models use a bidirectional encoder. Both encoder and decoder LSTMs have 512 hidden units, per gate layer normalization~\cite{Ba2016}, and residual skip connections after the second layer. 
The models are regularized using dropout with probability 0.2 and label smoothing with an uncertainty of 0.1~\cite{Szegedy2016}. Models are optimized until convergence using data parallelism over 32 P100s, using Adam~\cite{Kingma2015} with 
the learning rate schedule described in \newcite{Chen2018} and a batch size of 4,096 sentence-pairs per GPU. 
Checkpoints are selected based on development loss.
All streaming models use greedy decoding,
while offline models use beam search with a beam size of 20.

We implement soft attention, monotonic attention, MoChA, MILk and wait-$k$ as instantiations of an attention interface in a common code base, allowing us to isolate their contributions.
By analyzing development sentence lengths, we determined that wait-$k$ should employ a emission rate of 1 for DeEn, and 1.1 for EnFr.

\subsection{Development}
\begin{table}[t]
    \centering
    \begin{tabular}{c|rr|rr}
         & \multicolumn{2}{c}{unpreserved} & \multicolumn{2}{|c}{preserved} \\
        $\lambda$ & BLEU & DAL & BLEU & DAL\\ \hline
        0.0 & 27.7 & 21.0 & 27.7 & 27.9\\
        0.1 & 27.0 & 13.6 & 27.6 & 10.5\\
        0.2 & 25.7 & 11.6 & 27.5 & 8.7\\
    \end{tabular}
    \caption{Varying MILk's $\lambda$ with and without mass preservation on the DeEn development set.}
    \label{tab:dev_residual_fix}
\end{table}

We tuned MILk on our DeEn development set.
Two factors were crucial for good performance:
the preservation of monotonic mass (Section~\ref{sec:residual_fix}),
and the proper tuning of the noise parameter $n$ in Equation~\ref{eq:psn}, which
controls the discreteness of monotonic attention probabilities during training.

Table~\ref{tab:dev_residual_fix} contrasts MILk's best configuration before mass 
preservation against our final system.
Before preservation, MILk with a latency weight $\lambda=0$ still showed a substantial reduction in latency from 
the maximum value of 27.9, indicating an intrinsic latency incentive.
Furthermore, training quickly destabilized, resulting in very poor trade-offs for
$\lambda$s as low as $0.2$.

\begin{table}[t]
    \centering
    \begin{tabular}{c|rr}
        $n$ & BLEU & DAL  \\ \hline
        0 & 3.4 & 24.2\\
        1 & 10.8 & 12.9\\
        2 & 24.6 & 12.3\\
        3 & 27.5 & 10.4\\
        \textbf{4} & \textbf{27.5} & \textbf{8.7}\\
        6 & 26.3 & 7.2\\
    \end{tabular}
    \caption{Varying MILk's discreteness parameter $n$ with $\lambda$ fixed at 0.2 on 
    the DeEn development set.}
    \label{tab:dev_noise}
\end{table}

After modifying MILk to preserve mass, we then optimized noise with 
$\lambda$ fixed at a low but relevant value of 0.2,
as shown in Table~\ref{tab:dev_noise}.
We then proceeded the deploy the selected value of $n=4$ for testing both DeEn and EnFr.

\subsection{Comparison with the state-of-the-art}
\label{sec:exp_compare}
We compare MILk to wait-k, the current state-of-the-art in simultaneous NMT.
We also include MILk's predecessors, Monotonic Attention and MoChA, which
have not previously been evaluated with latency metrics.
We plot latency-quality curves for each system,
reporting quality using BLEU, and latency using Differentiable Average Lagging (DAL), 
Average Lagging (AL) or Average Proportion (AP) (see Section~\ref{sec:IAL}). 
We focus our analysis on DAL unless stated otherwise.
MILk curves are produced by varying the latency loss weight $\lambda$,\footnote{%
$\lambda=0.75$, 0.5, 0.4, 0.3, 0.2, 0.1, 0.05, 0.01, 0.0
}
wait-$k$ curves by varying $k$,\footnote{%
$k=2, 4, 6, 8, 10, 12, 14, 16, 20, 24, 300$}
and MoChA curves by varying chunk size.\footnote{%
$\mathit{cs}=1$ (Monotonic Attention), 2, 4, 8, and 16}
Both MILk and wait-$k$ have settings ($\lambda=0$ and $k=300$) corresponding
to full attention. 

\begin{figure}[t]
    \centering
        \includegraphics[width=0.48\textwidth]{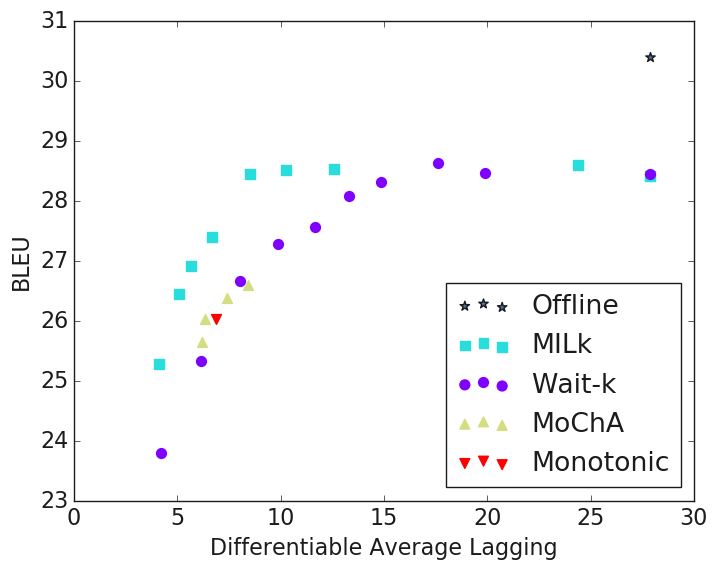}
        \includegraphics[width=0.23\textwidth]{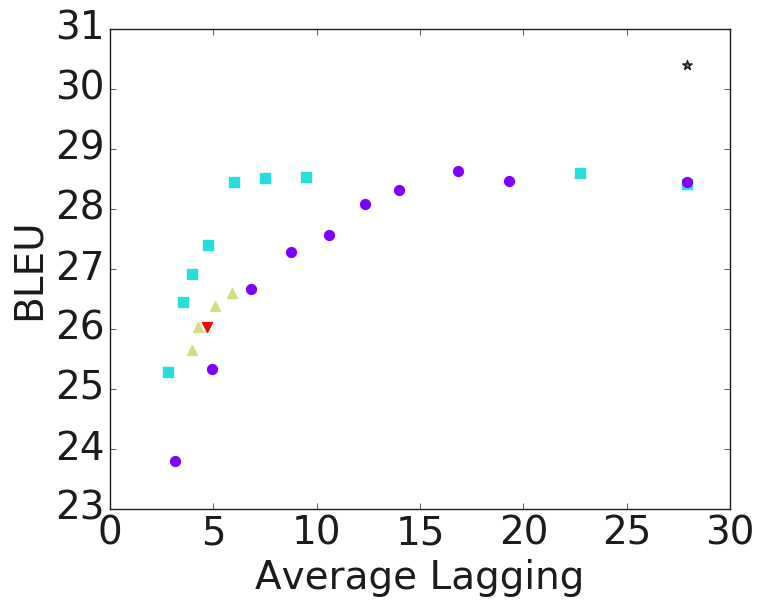}
        \includegraphics[width=0.23\textwidth]{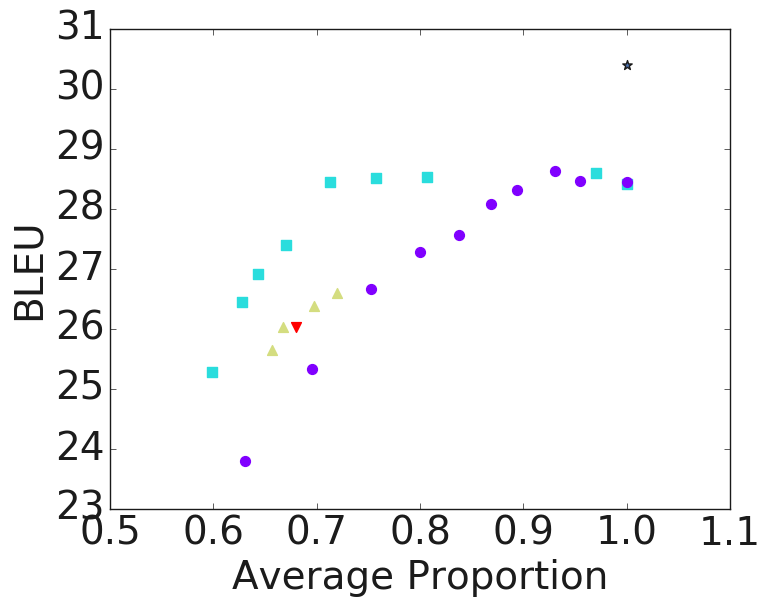}
        \caption{Quality-latency comparison for German-to-English WMT15 (DeEn) with
        DAL (upper), AL (lower-left), AP (lower-right).} 
        \label{fig:DeEn_compare}
\end{figure}

\begin{figure}[t]
    \centering
        \includegraphics[width=0.48\textwidth]{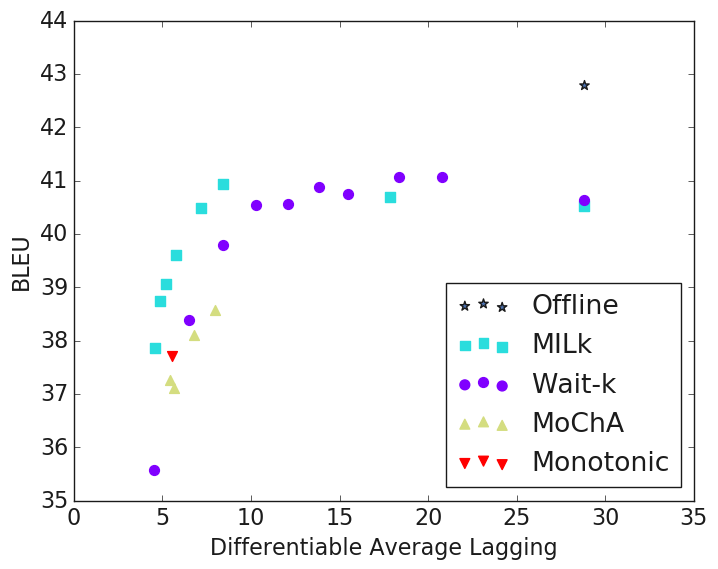}
        \includegraphics[width=0.23\textwidth]{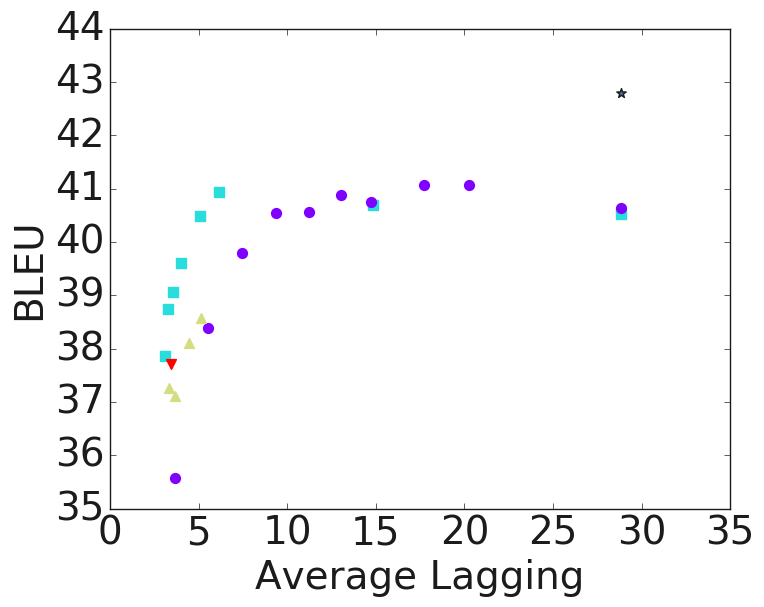} \includegraphics[width=0.23\textwidth]{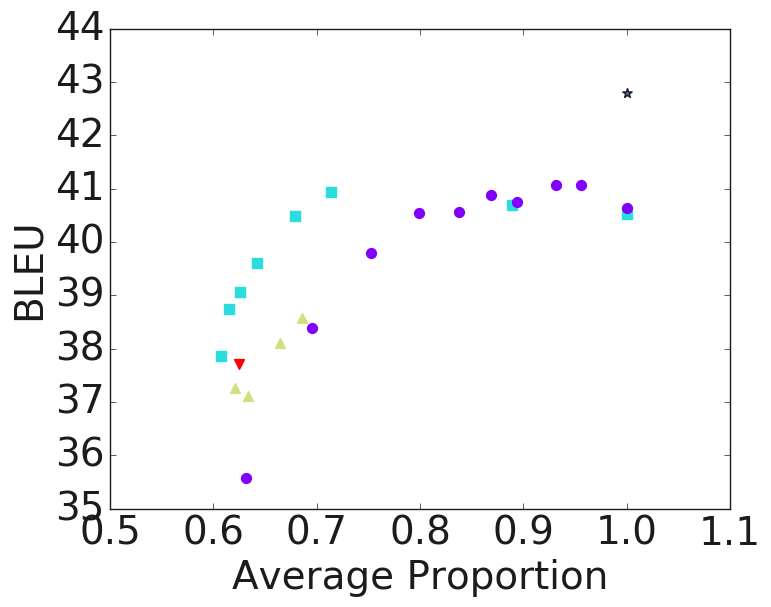}
        \caption{Quality-latency comparison for English-to-French WMT14 (EnFr) with
        DAL (upper), AL (lower-left), AP (lower-right).} 
        \label{fig:EnFr_compare}
\end{figure}

Results are shown in Figures~\ref{fig:DeEn_compare} and \ref{fig:EnFr_compare}.%
\footnote{%
Full sized graphs for all latency metrics, along with the corresponding numeric scores 
are available in Appendix A, included as supplementary material.
}
For DeEn, we begin by noting that
MILk has a clear separation above its predecessors MoChA and Monotonic Attention,
indicating that the infinite lookback is indeed a better fit for translation.
Furthermore, MILk is consistently above wait-$k$ for lags between 4 and 14 tokens.
MILk is able to retain the quality of full attention (28.4 BLEU) up to a lag of 8.5 
tokens, while wait-$k$ begins to fall off for lags below  13.3 tokens.
At the lowest comparable latency (4 tokens), MILk is 1.5 BLEU points ahead of wait-$k$.

EnFr is a much easier language pair: both MILk and wait-$k$ maintain the BLEU of full attention at lags of 10 tokens.
However, we were surprised to see that this does not mean we can
safely deploy very low $k$s for wait-$k$; its
quality drops off surprisingly quickly at $k=8$ (DAL=8.4, BLEU=39.8).
MILk extends the flat ``safe'' region of the curve out to a lag of 7.2 (BLEU=40.5).
At the lowest comparable lag (4.5 tokens), MILk once again surpasses wait-$k$, 
this time by 2.3 BLEU points.

The $k=2$ point for wait-$k$ has been omitted from all graphs to
improve clarity.
The omitted BLEU/DAL pairs are 19.5/2.5 for DeEn and 28.9/2.9 for EnFr,
both of which trade very large losses in BLEU for small gains in 
lag.
However, wait-$k$'s ability to function at all at such low latencies is notable.
The configuration of MILk tested here was unable to drop below lags of 4.

Despite MILk having been optimized for DAL, 
MILk's separation above wait-$k$ only grows
as we move to the more established metrics AL and AP.
DAL's minimum delay for each target token makes it far more conservative
than AL or AP.
Unlike DAL, these metrics reward MILk and its predecessors for their tendency
to make many consecutive writes in the middle of a sentence.

\subsection{Characterizing MILK's schedule}
\begin{figure}[t]
    \centering
    \includegraphics[width=0.35\textwidth]{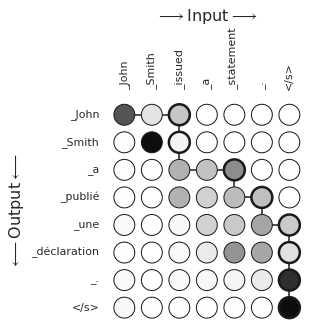}
    \includegraphics[width=0.4\textwidth]{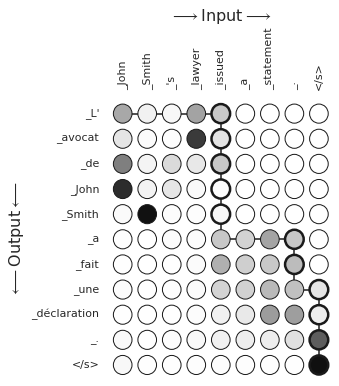}
    \caption{Two EnFr sentences constructed to contrast MILk's handling of a short noun phrase \ex{John Smith} against the longer \ex{John Smith's lawyer}. Translated by MILk with $\lambda=0.2$.}
    \label{fig:EnFr_milk_Smith}
\end{figure}
We begin with a qualitative characterization of MILk's behavior
by providing diagrams of MILk's attention distributions.
The shade of each circle indicates the strength of the soft alignment, while bold outlines indicate the location of the hard attention head,
whose movement is tracked by connecting lines.

In general, the attention head seems to loosely follow noun- and 
verb-phrase boundaries,
reading one or two tokens past the end of the phrase to ensure it is 
complete.
This behavior and its benefits are shown in 
Figure~\ref{fig:EnFr_milk_Smith},
which contrast the simple noun phrase \ex{John Smith} against
the more complex \ex{John Smith's laywer}.
By waiting until the end of both phrases, MILk is able to correctly 
re-order \ex{avocat} (\ex{lawyer}).

\begin{figure}[t]
    \centering
    \includegraphics[width=0.48\textwidth]{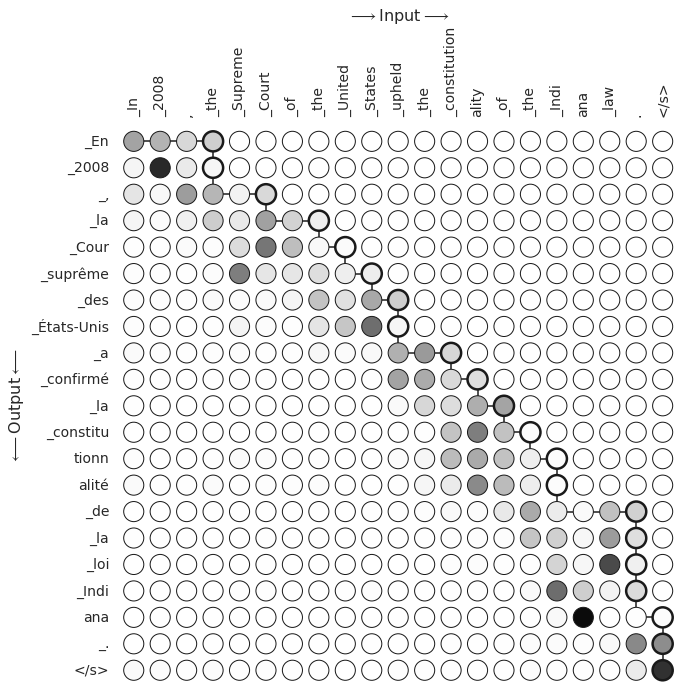}
    \caption{An example EnFr sentence drawn from our development set, as translated by MILk with $\lambda=0.2$.}
    \label{fig:EnFr_milk_supreme}
\end{figure}
\begin{figure}[t]
    \centering
    \includegraphics[width=0.48\textwidth]{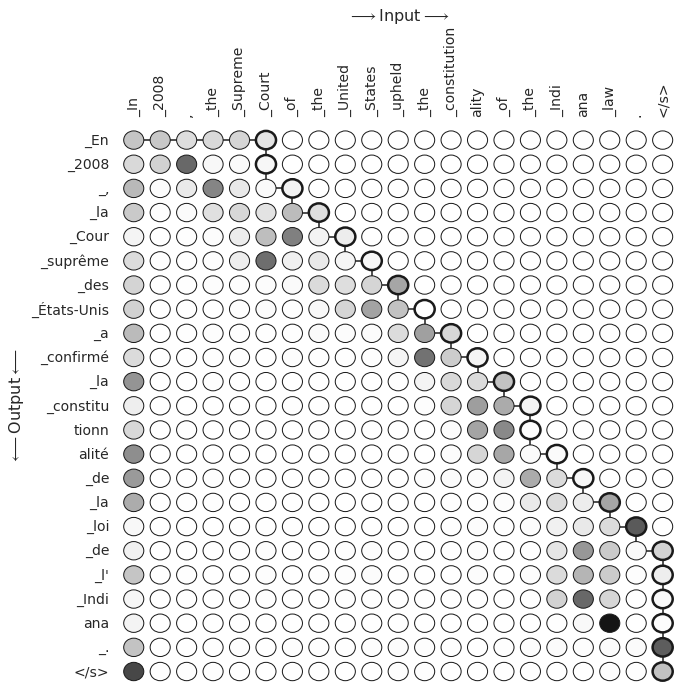}
    \caption{An example EnFr sentence drawn from our development set, as translated by wait-6.}
    \label{fig:EnFr_waitk_supreme}
\end{figure}

Figure~\ref{fig:EnFr_milk_supreme} shows a more complex sentence
drawn from our development set.
MILk gets going after reading just 4 tokens, writing the relatively safe, \ex{En 2008}.
It does wait, but it saves its pauses for tokens with likely future dependencies.
A particularly interesting pause occurs before the \ex{de} in \ex{de la loi}.
This preposition could be either \ex{de la} or \ex{du}, depending on the phrase it modifies.
We can see MILk pause long enough to read one token after \ex{law},
allowing it to correctly choose \ex{de la} to match the feminine \ex{loi} (\ex{law}).

Looking at the corresponding wait-6 run in Figure~\ref{fig:EnFr_waitk_supreme}, we can see that wait-6's fixed
schedule does not read \ex{law} before writing the same \ex{de}.
To its credit, wait-6 anticipates correctly, also choosing \ex{de la}, likely due to the legal context provided by the nearby phrase, \ex{the constitutionality}.

\begin{figure}[t]
    \centering
    \includegraphics[width=0.4\textwidth]{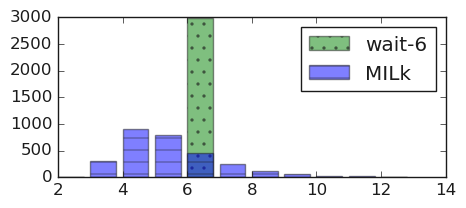}
    \caption{Histogram of initial delays for MILk ($\lambda=0.2$) and wait-6 on the EnFr development set.}
    \label{fig:EnFr_hist}
\end{figure}

We can also perform a  quantitative analysis of MILk's adaptivity by monitoring its initial delays;
that is, how many source tokens does it read before writing its first target token?
We decode our EnFr development set with MILk $\lambda=0.2$ as well as wait-6 and count the initial delays for each.\footnote{Wait-6 will have delays 
different from 6 only for source sentences with fewer than 6 tokens.}
The resulting histogram is shown in Figure~\ref{fig:EnFr_hist}.
We can see that MILk has a lot of variance in its initial delays,
especially when compared to the near-static wait-6.
This is despite them having very similar DALs: 5.8 for MILk and 6.5 for wait-6.

\section{Conclusion}
We have presented Monotonic Infinite Lookback (MILk) attention, 
an attention mechanism that uses a hard, monotonic head to manage the reading
of the source, and a soft traditional head to attend over whatever has been 
read.
This allowed us to build a simultaneous NMT system that is trained jointly with 
its adaptive schedule.
Along the way, we contributed latency-augmented training and a differentiable 
latency metric. 
We have shown MILk to have favorable quality-latency trade-offs compared to both
wait-$k$ and to earlier monotonic attention mechanisms.
It is particularly useful for extending the length of the region on the latency
curve where we do not yet incur a major reduction in BLEU.

\bibliography{acl2019}
\bibliographystyle{acl_natbib}

\appendix

\section{Expanded Results}
\begin{figure*}[p]
    \centering
    \begin{subfigure}[b]{0.48\textwidth}
        \includegraphics[width=\textwidth]{DeEn_DAL.png}
        \includegraphics[width=\textwidth]{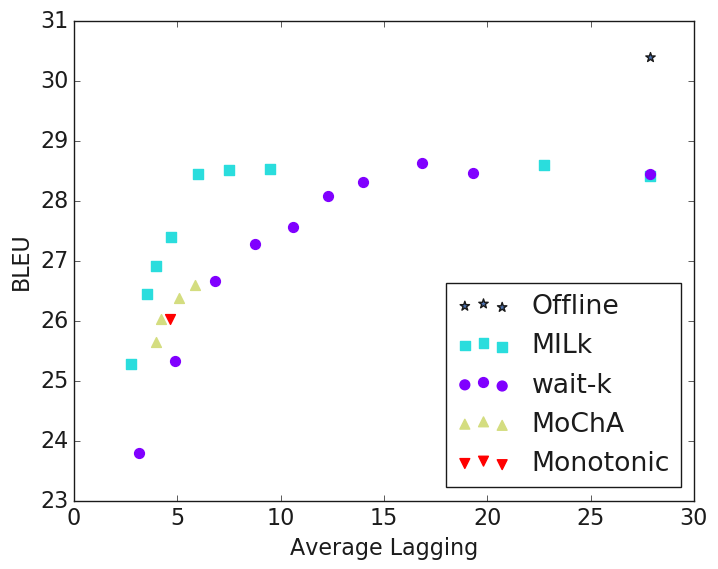}
        \includegraphics[width=\textwidth]{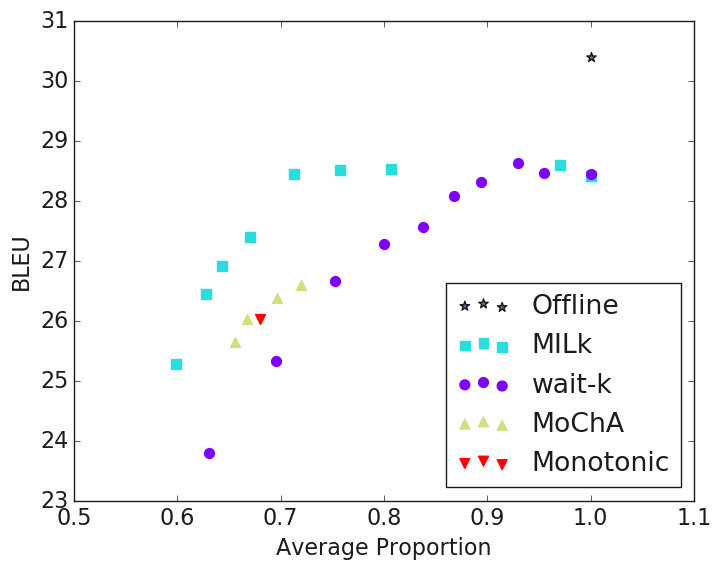}
        \caption{German-to-English WMT15 (DeEn) test set.} 
        \label{fig:DeEn_compare}
    \end{subfigure}
    \begin{subfigure}[b]{0.48\textwidth}
        \includegraphics[width=\textwidth]{EnFr_DAL.png}
        \includegraphics[width=\textwidth]{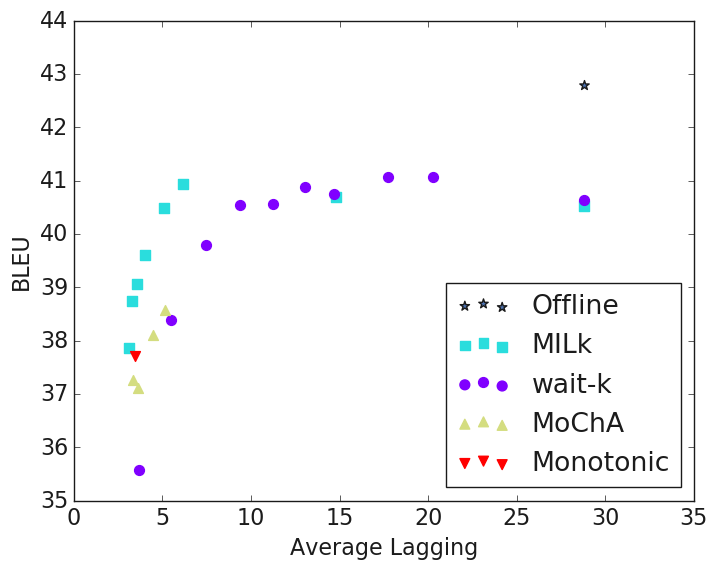}
        \includegraphics[width=\textwidth]{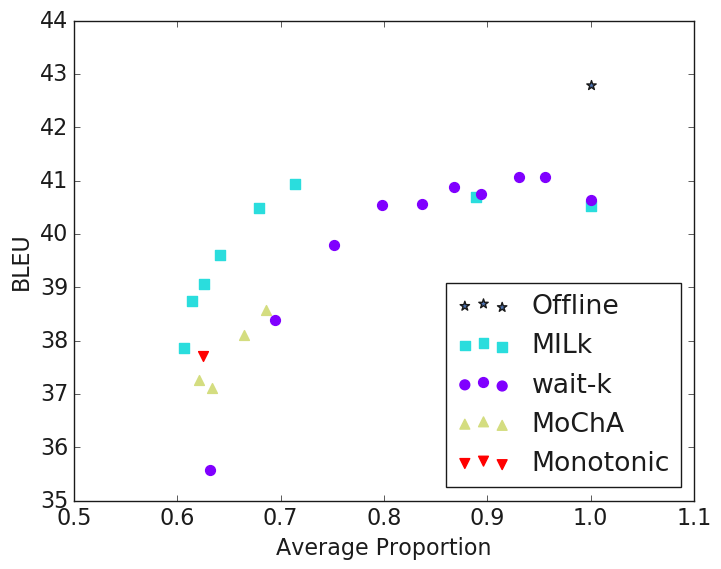}
        \caption{English-to-French WMT14 (EnFr) test set.} 
        \label{fig:EnFr_compare}
    \end{subfigure}
    \caption{BLEU versus latency (top: differentiable average lagging, middle: average lagging, bottom: average proportion) for
    our two language pairs (left: DeEn, right: EnFr).
    }
    \label{fig:all_compare}
\end{figure*}

We provide full-sized versions of our Quality-Latency curves from Section 5.3 in
Figure~\ref{fig:all_compare}. We also provide a complete table of results in Tables~\ref{tab:deen_all_compare} and \ref{tab:enfr_all_compare}.
As in the main text, DAL is Differentiable Average Lagging, AL is Average Lagging
and AP is Average Proportion.
wait-$k$ is parameterized by $k$, MoChA by its chunk size $\mathit{cs}$ and
MILk by its latency weight $\lambda$.
Results for EnFr MILk
with $\lambda=0.75$ are omitted, as it failed to converge.
\clearpage
\clearpage
\begin{table}[p]
    \centering
    \begin{tabular}{ll|rrrr}
    \multicolumn{2}{c}{Method} & BLEU & DAL & AL & AP  \\
    \hline
    \multirow{11}{*}{\rotatebox[origin=c]{90}{wait-$k$}}
     & $k=002$ & 19.5 & 2.5 & 1.5 & 0.56 \\
     & $k=004$ & 23.8 & 4.2 & 3.1 & 0.63 \\
     & $k=006$ & 25.3 & 6.1 & 4.9 & 0.70 \\
     & $k=008$ & 26.7 & 8.1 & 6.8 & 0.75 \\
     & $k=010$ & 27.3 & 9.9 & 8.8 & 0.80 \\
     & $k=012$ & 27.6 & 11.7 & 10.6 & 0.84 \\
     & $k=014$ & 28.1 & 13.3 & 12.3 & 0.87 \\
     & $k=016$ & 28.3 & 14.9 & 14.0 & 0.89 \\
     & $k=020$ & 28.6 & 17.6 & 16.9 & 0.93 \\
     & $k=024$ & 28.5 & 19.9 & 19.3 & 0.95 \\
     & $k=300$ & 28.4 & 27.9 & 27.9 & 1.00 \\
     \hline
     \multirow{5}{*}{\rotatebox[origin=c]{90}{MoChA}}
     & $\mathit{cs}=01$ & 26.0 & 6.9 & 4.7 & 0.68 \\
     & $\mathit{cs}=02$ & 25.6 & 6.2 & 4.0 & 0.66 \\
     & $\mathit{cs}=04$ & 26.0 & 6.4 & 4.2 & 0.67 \\
     & $\mathit{cs}=08$ & 26.4 & 7.4 & 5.1 & 0.70 \\
     & $\mathit{cs}=16$ & 26.6 & 8.4 & 5.9 & 0.72 \\
     \hline
     \multirow{9}{*}{\rotatebox[origin=c]{90}{MILk}}
     & $\lambda=0.75$ & 25.3 & 4.1 & 2.8 & 0.60 \\
     & $\lambda=0.50$ & 26.4 & 5.1 & 3.5 & 0.63 \\
     & $\lambda=0.40$ & 26.9 & 5.7 & 4.0 & 0.64 \\
     & $\lambda=0.30$ & 27.4 & 6.7 & 4.7 & 0.67 \\
     & $\lambda=0.20$ & 28.4 & 8.5 & 6.0 & 0.71 \\
     & $\lambda=0.10$ & 28.5 & 10.3 & 7.5 & 0.76 \\
     & $\lambda=0.05$ & 28.5 & 12.6 & 9.5 & 0.81 \\
     & $\lambda=0.01$ & 28.6 & 24.4 & 22.7 & 0.97 \\
     & $\lambda=0.00$ & 28.4 & 27.9 & 27.9 & 1.00 \\
    \end{tabular}
    \caption{Complete DeEn test set results, backing the curves in Figure~\ref{fig:DeEn_compare}.}
    \label{tab:deen_all_compare}
\end{table}

\begin{table}[p]
    \centering
    \begin{tabular}{ll|rrrr}
    \multicolumn{2}{c}{Method} & BLEU & DAL & AL & AP  \\
    \hline
    \multirow{11}{*}{\rotatebox[origin=c]{90}{wait-$k$}}
     & $k=002$ & 28.9 & 2.9 & 2.1 & 0.57 \\
     & $k=004$ & 35.6 & 4.5 & 3.7 & 0.63 \\
     & $k=006$ & 38.4 & 6.5 & 5.5 & 0.70 \\
     & $k=008$ & 39.8 & 8.4 & 7.5 & 0.75 \\
     & $k=010$ & 40.5 & 10.3 & 9.4 & 0.80 \\
     & $k=012$ & 40.6 & 12.1 & 11.3 & 0.84 \\
     & $k=014$ & 40.9 & 13.9 & 13.0 & 0.87 \\
     & $k=016$ & 40.7 & 15.5 & 14.7 & 0.89 \\
     & $k=020$ & 41.1 & 18.3 & 17.7 & 0.93 \\
     & $k=024$ & 41.1 & 20.8 & 20.3 & 0.96 \\
     & $k=300$ & 40.6 & 28.8 & 28.8 & 1.00 \\
     \hline
     \multirow{5}{*}{\rotatebox[origin=c]{90}{MoChA}}
     & $\mathit{cs}=01$ & 37.7 & 5.5 & 3.4 & 0.63 \\
     & $\mathit{cs}=02$ & 37.3 & 5.4 & 3.3 & 0.62 \\
     & $\mathit{cs}=04$ & 37.1 & 5.6 & 3.6 & 0.63 \\
     & $\mathit{cs}=08$ & 38.1 & 6.8 & 4.5 & 0.66 \\
     & $\mathit{cs}=16$ & 38.6 & 7.9 & 5.1 & 0.69 \\
     \hline
     \multirow{9}{*}{\rotatebox[origin=c]{90}{MILk}}
     & $\lambda=0.50$ & 37.9 & 4.6 & 3.1 & 0.61 \\
     & $\lambda=0.40$ & 38.7 & 4.9 & 3.3 & 0.61 \\
     & $\lambda=0.30$ & 39.1 & 5.2 & 3.6 & 0.63 \\
     & $\lambda=0.20$ & 39.6 & 5.8 & 4.0 & 0.64 \\
     & $\lambda=0.10$ & 40.5 & 7.2 & 5.1 & 0.68 \\
     & $\lambda=0.05$ & 40.9 & 8.4 & 6.2 & 0.71 \\
     & $\lambda=0.01$ & 40.7 & 17.9 & 14.8 & 0.89 \\
     & $\lambda=0.00$ & 40.5 & 28.8 & 28.8 & 1.00 \\
    \end{tabular}
    \caption{Complete EnFr test set results, backing the curves in Figure~\ref{fig:EnFr_compare}.}
    \label{tab:enfr_all_compare}
\end{table}

\end{document}